\documentclass{article}

\usepackage{arxiv}

\usepackage[utf8]{inputenc} 
\usepackage[T1]{fontenc}    
\usepackage{hyperref}       
\usepackage{url}            
\usepackage{booktabs}       
\usepackage{amsfonts}       
\usepackage{nicefrac}       
\usepackage{microtype}      
\usepackage{lipsum}		
\usepackage{graphicx}
\usepackage{natbib}
\usepackage{doi}

\usepackage{amsthm}

\usepackage{amsmath,amssymb,amsfonts}
\usepackage{enumitem}
\usepackage{import}

\usepackage{algorithm}
\usepackage{algpseudocode}
\algtext*{EndIf}                        
\algtext*{EndFor}                       
\algtext*{Function}                     
\algtext*{EndFunction}                  
\makeatletter
\renewcommand{\ALG@beginalgorithmic}{\small}
\makeatother
\usepackage{threeparttable}

\usepackage{textcomp}
\usepackage{xcolor}

\usepackage{hyperref} 
\usepackage[all]{hypcap}
\usepackage{url}
\usepackage[utf8]{inputenc}

\newlist{questions}{enumerate}{2}
\setlist[questions,1]{label=RQ\arabic*.,ref=RQ\arabic*}
\setlist[questions,2]{label=(\alph*),ref=\thequestionsi(\alph*)}

\usepackage{enumitem}
\usepackage{import}

\title{Towards a User Privacy-Aware Mobile Gaming App Installation Prediction Model}

\author{ 
\href{https://orcid.org/0000-0000-0000-0000}{\includegraphics[scale=0.06]{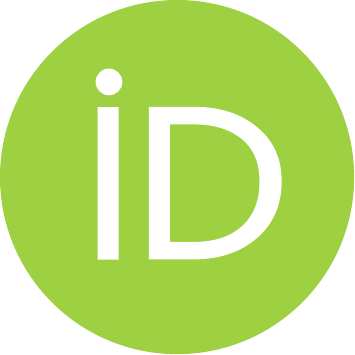}\hspace{1mm}Ido Zehori}\thanks{Ido Zehori is the corresponding author} \\
	Bigabid Inc.\\
	Tel-Aviv, Israel \\
	\texttt{Ido.z@bigabid.com} \\
 \And
 \href{https://orcid.org/0000-0002-8086-2246}{\includegraphics[scale=0.06]{orcid.pdf}\hspace{1mm}Nevo Itzhak} \\
	Software and Information Systems Engineering\\
	Ben-Gurion University of the Negev\\
	Beer-Sheva, 8410501, Negev, Israel \\
	\texttt{nevoit@post.bgu.ac.il} \\
	\And
	\href{https://orcid.org/0000-0003-0328-2333}{\includegraphics[scale=0.06]{orcid.pdf}\hspace{1mm}Yuval Shahar} \\
	Software and Information Systems Engineering\\
	Ben-Gurion University of the Negev\\
	Beer-Sheva, 8410501, Negev, Israel \\
	\texttt{yshahar@bgu.ac.il} \\
     \And
	\href{https://orcid.org/0000-0000-0000-0000}{\includegraphics[scale=0.06]{orcid.pdf}\hspace{1mm}Mia Dor Schiller} \\
	Bigabid Inc.\\
	Tel-Aviv, Israel \\
	\texttt{Mia.d@bigabid.com} \\
}

\renewcommand{\shorttitle}{Towards a User Privacy-Aware Installation Prediction Model}

\hypersetup{
pdftitle={Towards a User Privacy-Aware Installation Prediction Model},
pdfsubject={q-cs.LG},
pdfauthor={Ido Zehori, Nevo Itzhak, Yuval Shahar, Mia Dor Schiller},
pdfkeywords={demand-side platform, mobile gaming app, privacy-aware machine learning},
}

\begin{document}
\maketitle

\begin{abstract}
Over the past decade, programmatic advertising has received a great deal of attention in the online advertising industry.
A real-time bidding (RTB) system is rapidly becoming the most popular method to buy and sell online advertising impressions.
Within the RTB system, demand-side platforms (DSP) aim to spend advertisers' campaign budgets efficiently while maximizing profit, seeking impressions that result in high user responses, such as clicks or installs.

In the current study, we investigate the process of predicting a mobile gaming app installation from the point of view of a particular DSP, while paying attention to user privacy, and exploring the trade-off between privacy preservation and model performance. 
There are multiple levels of potential threats to user privacy, depending on the privacy leaks associated with the data-sharing process, such as data transformation or de-anonymization.

To address these concerns, privacy-preserving techniques were proposed, such as cryptographic approaches, for training privacy-aware machine-learning models.
However, the ability to train a mobile gaming app installation prediction model without using user-level data, can prevent these threats and protect the users' privacy, even though the model's ability to predict may be impaired.
Additionally, current laws might force companies to declare that they are collecting data, and might even give the user the option to opt out of such data collection, which might threaten companies’ business models in digital advertising, which are dependent on the collection and use of user-level data.

In this study, we show that learning a user privacy-aware mobile gaming app installation prediction model leads to an AUC-ROC of 81\% and an AUPRC of 49\% while using a non-privacy-aware model leads to an AUC-ROC of 89\% and AUPRC of 66\%.

We conclude that privacy-aware models might still preserve significant capabilities, enabling companies to make better decisions, dependent on the privacy-efficacy trade-off utility function of each case.
\end{abstract}


\newpage

\section{Introduction}
\label{sec:introduction}
\begin{figure}[!h]
    \centering
    \includegraphics[width=\linewidth]{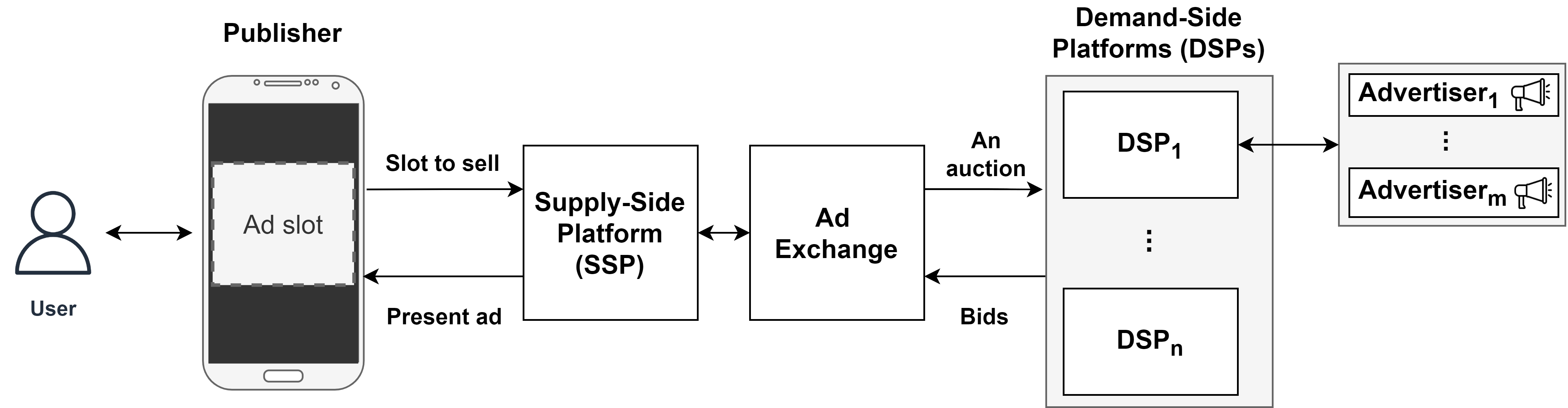}
    \caption{A high-level overview of the real-time bidding procedure. 
    A publisher offers an ad slot, the SSP provides information about the user and the publisher to the ad exchange system, and the DSPs participate in an auction.}
    \label{fig:datatypes}
\end{figure}

Programmatic advertising has received great attention in the online advertising industry in the last decade. 
Real-time bidding (RTB) is quickly becoming the leading method \citep{spencer2011arrival} to enable buying and selling online advertising impressions (i.e., single-user view of an advertisement) through real-time auctions.
The RTB display advertising (Fig. \ref{fig:datatypes}) is composed of three main components: (1) the demand-side platform (DSP) that enables advertisers to automate the purchasing of the advertising impressions; (2) the supply-side platform (SSP) that enables the publishers (e.g., application owners) to sell the advertising impression, and (3) advertising exchange system that connect advertisers and publishers.

As illustrated in Figure \ref{fig:datatypes}, given an advertising slot offered by the publisher to sell, the SSP provides impression information to an advertising exchange system, which creates an auction and forwards that information to the DSPs (e.g., Google Ads, Facebook Ads Manager, AppLovin, IronSource, Liftoff, or Moloco).
Once the auction is completed, the winner is chosen, and an advertisement is served on the publisher's website or application.
For example, suppose there are three auction participants (i.e., DSPs), $d_1$, $d_2$, and $d_3$, that compete on a specific advertising impression, and their offered prices, according to their secret bidding strategies, were \$1.3, \$1.5, and \$0.7, respectively. 
Then, the winning price is the highest of their offers, which was \$1.5, offered by DSP $d_2$.
In mobile performance advertising, a DSP's objectives extend beyond getting the application installation, but also the prediction of user engagement probabilities, whether through in-app purchases or watching advertisements.
It is worth noting that while some companies may own different parts of the chain in the real-time bidding procedure, transparency, and fairness are among the challenges in today's online advertising industry.
In light of the potential conflicts of interest created by this ownership structure, some players may have an unfair advantage.

From the DSP's perspective, the aim is to spend the advertisers' campaign budget effectively and to achieve high profit (i.e., increasing the profit margin).
That means a DSP seeks advertising impressions that result in high user responses, such as clicks or installs, but within budget constraints.
To achieve that, a DSP typically considers three aspects given an auction: utility, cost, and deciding on a bidding strategy.
The utility represents the value to the DSP of having the user click on the ad, while utility estimation represents the probability of a user being interested in the ads by clicking, installing the app, or depositing money into the app.
The cost refers to the estimated winning price of that auction.
Estimating the auction winning price is critical in designing bidding strategies; however, it is challenging.
In practice, a DSP can not observe the other participants' bids during the auction or when the auction is completed, but only the highest bid price when the auction is completed.
For RTB with significant volume, in which the number of bidders participating in the auction is potentially high, the average winning rate is naturally lower for each participant.
The last aspect is a bidding strategy that incorporates that estimated utility and winning price to decide on the offered price in an auction (if at all).

In today's digital world, data have become a valuable asset for companies, and it is important for organizations to protect these data to maintain the trust of their customers.
Data-driven advertising platforms, such as DSPs, often collect and store user-level information for research purposes, including sensitive data such as financial information.
As a result, it is crucial for DSPs to adhere to strict privacy standards and implement robust security measures to protect these data.
Multiple levels of security threats are associated with data sharing, including data transformations and deanonymization.
Machine learning models that are privacy-aware were developed using privacy-preserving techniques, such as cryptographic approaches.
\newpage \noindent
The methods still have some limitations, such as flexibility to cope with ongoing advances in machine learning, or scalability as they require additional processing and communication costs \citep{al2019privacy}.
In this regard, the ability to learn a machine learning model without involving user-level data can both prevent privacy and security-related threats, even though the model's prediction capability may be impaired.
Additionally, more and more companies, including Big Tech firms, are changing their policies to put user privacy at the forefront.
For Apple’s iOS mobile operating systems, cross-site identity matching was restricted in April 2021. 
Thus, iOS applications must receive explicit consent before exchanging user-level data with third parties. 
Government regulations play an important role in these changes as well, with regulations such as the General Data Protection Regulation (GDPR) \cite{hoofnagle2019european} and the California Consumer Privacy Act (CCPA) \cite{pardau2018california} imposing strict requirements on user data collection, storage, and use.
As users' consent rates are low \citep{wagner_2021}, such restriction might threaten companies that are strongly dependent on the collection and use of user-level data \citep{morrison_2021}, such as in online advertising markets.

In this study, we investigate the utility modeling and prediction of mobile app installation from the point of view of a particular DSP.
We compare the prediction performance of the mobile gaming app installation prediction models with and without different feature sets, such as the user or publisher-level data, and their combinations.
We consider also an important aspect that is typically ignored in previous studies, namely, that the bidding and winning processes are executed when a user loads a page, which requires quick decision-making from the DSP regarding the bidding.
Most previous studies evaluated their methods on publicly available RTB datasets iPinYou \citep{liao2014ipinyou}, and YOYI \citep{ren2017bidding}.
However, in this paper, we evaluated our methods on a new dataset called The Bigabid dataset (Sec. \ref{sec:datasets}), which contains data from the real-world mobile gaming advertising industry.

The main contribution of the paper is the investigation of the trade-off between users' privacy and the prediction power of using user-level data for predicting mobile gaming application installation in digital advertising.
Our study demonstrates that by considering user privacy in the mobile gaming app installation prediction model, the performance of the model is less accurate, with an AUC-ROC of 81\% and an AUPRC of 49\%, however, when user privacy is not taken into account, the model's performance is better with an AUC-ROC of 89\% and an AUPRC of 66\%.
These surprising findings showed that, despite lacking user-level data, our models still managed to exhibit good performance in predicting app installations.

   
    

The rest of this paper is organized as follows: we start by reviewing the literature (Sec.~\ref{sec:background}), considering the RTB aspects - utility, cost, and bidding strategy, and then, the topic of privacy-aware machine learning.
We then proceed with the Problem Definition (Sec.~\ref{sec:prob_statement}), Dataset (Sec.~\ref{sec:datasets}), Methods (Sec.~\ref{sec:methods}), Evaluation and Results (Sec.~\ref{sec:evaluation}--\ref{sec:results}), and finally summarize the work, present conclusions, and discuss future work (Sec.~\ref{sec:disc_and_conc}).

\section{Related Work}
\label{sec:background}

\subsection{Utility Estimation}
\label{sec:Utility_Estimation}

Compared to traditional offline advertising, online advertising offers the ability to target individual users by user response (e.g., clicks, conversions) prediction.
Advertisers usually measure the effectiveness of their ads by measuring user responses and makes payment only when these actions are triggered by their advertisement.
Creating effective advertising requires DSPs to precisely predict users' responses, since accurate predictions could greatly improve the subsequent decision-making process, such as bid optimization. 
Predicting whether a particular user will click a specific ad in a specific context is the most fundamental prediction task, which is referred to as click-through rate (CTR) prediction.

On large-scale advertising setups, logistic regression models have proven successful in predicting clicks \citep{richardson2007predicting,mcmahan2013ad} and conversions \citep{bhamidipati2017large}.
Although linear models are easy to implement, they do not learn the complex patterns to capture interactions between the assumed conditionally independent raw features, typically leading to low performance.

The use of nonlinear models may improve estimation performance because they can make use of different feature combinations.
One popular approach is to use deep learning models, which have achieved state-of-the-art performance in many prediction tasks, including CTR prediction \citep{zhou2018deep}.
Deep learning models can effectively capture complex patterns in high-dimensional data, which are often encountered in online advertising.
Previous studies suggested using historical data to estimate the expected value of a particular user's response.
For example, Qu et al. (\citeyear{qu2019dynamic}) proposed a method that uses the user's historical click and conversion data to estimate the expected value of a user's response.
However, this approach relies on the assumption that the user's behavior remains consistent over time, which may not always be the case.


\newpage

\subsection{Cost Prediction}
\label{sec:Cost_Prediction}

Predicting the winning price in action is an essential task since the auction participants (i.e., the DSPs) can use the predicted highest bid price to compute the winning rate of a bid and decide on their final bid price.


The cost predicting aspect is typically defined, rather than predicting the winning bid price, as a process of forecasting the market price distribution for auctions of a specific ad, named \textit{bid landscape forecasting}.
The estimated market price distribution and its cumulative distribution function allow computing the winning probability given each bid price. 

Learning the market price distribution plays a critical role in designing bidding strategies; however, it is challenging.
A DSP can not observe the other participants' bids during the auction or when the auction is completed, but only the highest bid price when the auction is completed.
Additionally, it becomes even more challenging to forecast the market price distribution according to its dynamic nature; different DSPs probably have different strategies and work with multiple advertisers, which can behave differently depending on the given bid.
These challenges become more noticeable for RTB with significant volume, in which the number of bidders participating in the auction is potentially high, which naturally lowers the winning rate for each participant.

Most recent studies considered in their methodology the second-price auction mechanism, in which the winning DSP is charged the price offered by the second-highest DSP.
The second-price auction mechanism brings significant bias to the bid landscape forecasting and requires using models that consider censored information.
However, since this auction mechanism is irrelevant to our dataset, in which the highest price wins the bid, the second-price auction mechanism will be ignored.



A regression model returns the excepted mean value for the winning price given a bid while ignoring the other auction participants and their strategies.
However, the winning price is dynamic and depends on the other DSPs' strategies.
Therefore, a dynamic model considers the negative examples, and the other DSPs' strategies should be learned.

Based on empirical analysis of real-world datasets, researchers proposed different function forms to model the market price distribution, such as the log-normal \citep{wu2018deep,cui2011bid}, Gaussian \citep{wu2015predicting}, and Gamma \citep{zhu2017gamma} distribution.
Then, the distribution parameters are found by maximizing the distribution's likelihood function.
Wu et al. \citep{wu2015predicting} combined a linear regression model and a censored linear regression model \citep{tobin1958estimation} for winning price estimation.
Using linear regression, assume the dependant variable (i.e., the winning price) is Gaussian distributed, and as a result, the values range from $(-\infty, +\infty)$.
As a result, the Gaussian distribution is unsuitable for modeling the winning price \citep{wu2018deep,zhu2017gamma}, which should only be positive values.
Gamma is a more suitable distribution for the winning price, as the values range from $(0, +\infty)$, as proposed by Zhu et al. \citep{zhu2017gamma}.
Yet, modeling the winning price by assuming a specific distribution is a strong assumption.
Thus, Wu et al. \citep{wu2018deep} proposed combining different winning price distributions to the deep learning models with censoring information to improve the model's flexibility.

Non-parametric distributions were also considered.
Wang et al. \citep{marmer2013model} clustered the feature vectors using decision trees and then used survival analysis with non-parametric distribution to model the winning price of each cluster.

\subsection{Bidding Strategy Optimization}
\label{sec:Bidding_Strategy_Optimization}
In RTB, ads are auctioned off in real-time to the highest bidder, making it essential for DSPs to have a strong bidding strategy in order to be successful.
One important aspect of bidding strategy optimization in RTB is the use of algorithms to make real-time decisions about bid prices \citep{liu2022real}.
By using these algorithms, DSPs can quickly and accurately determine the optimal bid price for each auction, taking into account a range of factors such as the expected impressions against the bid price (or bid landscape), total budget, and the campaign’s remaining lifetime.

Bidding strategies can be divided into two main categories: static and dynamic.
A static bidding strategy uses pre-determined formulas, such as linear \citep{perlich2012bid} or non-linear \citep{zhang2014optimal}, which are based on the probability of winning the auction and the prior distribution of the impression features. 
The simplicity and ease of deployment of static bidding strategies are the key reasons they are so widely adopted on DSPs.

However, these static bidding strategies are not effective in the highly dynamic nature of RTB auctions. Reinforcement learning-based bidding strategies have shown some promise in this area, such as RLB \citep{cai2017real}, DRLB \citep{wu2018budget}, and FAB \citep{liu2020dynamic}. A recent empirical analysis \citep{liu2022real} has shown that a DRLB-based framework for optimizing DSPs or advertisers' real-time bidding strategies is a promising solution.
However, reinforcement learning-based bidding strategies are not yet ready for commercial deployment.

\subsection{Privacy-Aware Machine Learning}
\label{sec:Privacy_Preserving_ML}
Privacy-aware machine learning is a subfield of machine learning that focuses on developing algorithms and techniques that can learn from data while preserving the privacy of individuals \citep{chaudhuri2011differentially}. There has been a growing concern in recent years about the potential risks to privacy posed by machine learning algorithms \citep{shokri2015privacy}, which are often able to extract sensitive information from data sets. 

One approach to privacy-aware machine learning is the use of privacy-preserving data perturbation techniques, which add noise to the data to prevent sensitive information from being revealed \citep{wood2018differential}. Another approach is the use of homomorphic encryption \citep{mohassel2017secureml,aono2017privacy}, which allows data to be encrypted in such a way that it can be used for machine learning tasks without revealing its original form.

In addition to these technical approaches, there has also been a focus on developing frameworks and ethical guidelines for privacy-aware machine learning \citep{fat2018fairness}.


\section{Problem Definition}
\label{sec:prob_statement}
Given a user $u$ visits the publisher $s$'s application, the real-time bidding (RTB) display advertising system offers advertising slot $ad$ to show to $u$ on $s$, the supply-side platform (SSP) provides $u$ and $p$'s information to $n$ different demand-side platforms (DSPs) $SP=\{{d}_1, {d}_2, ..., {d}_n \}$ throughout an auction $c$ that created by the advertising exchange system (Fig. \ref{fig:datatypes}).
From a specific DSP's $d \in SP$ perspective, given $c$ and $m$ different advertisers $R=\{{r}_1, {r}_2, ..., {r}_m \}$, $d$ aims of selecting $r\in R$ that will result in high profit by considering $u$'s probability of clicking and installing (utility; Sec. \ref{sec:Utility_Estimation}), estimated $c$'s winning price (cost; Sec. \ref{sec:Cost_Prediction}), and $r$'s campaign budget to decide on the offered price in $c$ (bidding strategy optimization; Sec. \ref{sec:Bidding_Strategy_Optimization}).
Once the auction $c$ is completed, given bidding prices of the $n$ DSPs $B=\{{b}_1, {b}_2, ..., {b}_n \}$, the $i$-th DSP $d_i \in SP$ that offered $b_i \in B$ wins $c$ if $b_i = max(B)$ and as a result, $r$'s advertisement is presented to $u$ on $p$.

We frame the problem of predicting whether $u$ will click and then install the mobile gaming application after watching the advertisement of $r \in R$ located at $ad$ on $s$ as a classification problem.
The goal is to predict the probability that the user will watch the mobile gaming application advertisement, click, and install the mobile gaming application, given $d \in SP$ wins $c$ and present $r \in R$'s advertisement.
Given a database $D^{w \times f}$ that includes records of winning auctions of a single DSP, and a labels vector $y^w$, where $w$ represents the number of auctions in the dataset and $f$ represents the number of attributes.
The label vector $y$ indicates whether the user clicked the mobile gaming app's advertisement and then install the mobile application.




\section{The Bigabid Dataset}
\label{sec:datasets}
The data for this study were collected from real-mobile gaming advertising campaigns run through the Bigabid DSP platform.
Bigabid is a leading DSP that specializes in the gaming app market. 
Bigabid operates at a high speed, bidding on over one million opportunities per second. Bigabid employs advanced machine learning techniques to train its business models on a vast dataset of over 4 billion unique devices.

\subsection{Dataset Creation}
The data were collected over a two-day period in January 2021. 
The data include 44,824 mobile gaming app impressions from 227 different mobile apps (i.e., publishers) that resulted in 10,000 clicks, and 7,872 app installations.
The sampling methodology employed for the dataset creation is designed to obfuscate the actual ratios.
The data include numeric variables only.

Additionally, since a DSP is unable to observe the outcome of auctions it did not win (e.g., clicks or installs), the data are limited to the auctions that Bigabid was successful in winning.
The data include all the clicks and installs that occurred during that period.
Even though the outcome of all auctions is not known, the distribution of the auctions that Bigabid won is similar to the real-world distribution. This is because Bigabid's bidding strategy considers the estimated utility and cost of each auction before deciding whether to bid on it. As a result, Bigabid may win auctions with high or low estimated utilities. This is a common challenge in RTB from the perspective of a DSP company.
To address this issue, Bigabid runs exploration campaigns to gather data that covers the entire feature space. 
This allows Bigabid to thoroughly investigate the underlying relationships and patterns within the data and scale the winning auctions data accordingly.
This helps Bigabid better predict auction outcomes and improve the performance of its auction platform on real-world distribution.
\newpage \noindent
However, as mentioned in the study limitations, this is not the exact distribution in the real world, as there can be auctions with outliers utility estimation.
To accurately learn and evaluate the real-world distribution, we would need to have all labels and predict from the perspective of the auction organizer.


\subsection{Data Structure}
The data can be divided into three main categories: user information, publisher, and meta-features:
\begin{enumerate}
    \item \textbf{User features} are characteristics of the users that do not change frequently and reflect their preferences and "static" behavior.  
    \item \textbf{Publisher features} are first-party data about the publisher in which Bigabid is presenting the ad, such as performance metrics and the type of publisher that is categorized according to Bigabid's deep categories system \citep{raz_2021}.
    \item \textbf{Meta features} include details about the specific ad bid, including the ad slot and the type of ad being presented. The meta-features help us to target our ads more effectively and improve the performance of our campaigns.
\end{enumerate}

\section{Methods}
\label{sec:methods}
XGBoost (eXtreme Gradient Boosting) \citep{chen2016xgboost} and LightGBM (Light Gradient Boosting Machine) \citep{ke2017lightgbm} are both gradient-boosting frameworks that were used in this study.
Gradient boosting means they use decision trees as weak learners and optimize the model using gradient descent.
Gradient boosting is an iterative process, where each iteration builds a new model that corrects the errors of the previous model. This allows XGBoost and LightGBM to achieve high levels of accuracy, even with complex and nonlinear data.

In gradient boosting for classification, the final classifier is a weighted sum of the individual weak learners, where the weights are determined by solving the following optimization problem:

$$F(x) = \arg\min_{F} \sum_{i=1}^n L(y_i, F(x_i)) + \sum_{j=1}^T \nu_j f_j(x)$$

where $L$ is a loss function, $T$ is the number of weak learners, and each $\nu_j$ the weight assigned to the weak learner. The loss function $L$ measures the difference between the predicted class label $F(x_i)$ and the true class label $y_i$. The first term in the formula, $\sum_{i=1}^n L(y_i, F(x_i))$, is the loss function, which measures the difference between the predicted class labels $F(x_i)$ and the true class labels $y_i$. The second term, $\sum_{j=1}^T \nu_j f_j(x)$, is the weighted sum of the individual weak learners, where $\nu_j$ are the weights assigned to each weak learner.

No imputation technique was used in this study as both XGBoost and LightGBM are able to handle missing values, which are common challenges in mobile app installation prediction.
XGBoost and LightGBM use advanced techniques such as regularization and feature selection to handle missing values and reduce overfitting.

There are some key differences between the two algorithms. XGBoost uses a more regularized model formalization, with an L2 regularization term on the weights of the trees to control overfitting, whereas LightGBM uses a novel gradient-based one-side sampling algorithm to speed up training.
Additionally, LightGBM uses histogram-based algorithms for decision tree learning, which can reduce the calculation time compared to XGBoost.

To find the optimal hyperparameters for our XGBoost and LightGBM models, we performed a grid search over the learning rate $\eta \in \{ 0.001, 0.01, 0.1 \}$, number of estimators $n_\text{estimators} \in \{ 10, 25, 50 \}$, and the maximum number of leaves in the trees $l_\text{max} \in \{ 7, 10, 15 \}$. The learning rate determines the step size at each iteration of the gradient descent algorithm, and the number of estimators and maximum number of leaves control the complexity of the models \citep{hastie2009elements}.



\newpage

\section{Evaluation}
\label{sec:evaluation}
We performed an experimental study to evaluate our method and compare its performance, with and without a different set of features, to baseline models.
The evaluation was performed on the Bigabid dataset, to answer three research questions that were defined.

\noindent The main \textit{research questions} for this study were: 

\begin{questions}[leftmargin=1.5cm]
    \item Which features set achieves the best prediction performance?
    \label{itm:features_set}
    \item Which machine learning model (XGBoost or LightGBM) achieves the best performance, in terms of prediction accuracy?
    \label{itm:model}
\end{questions}

\subsection{Experiments}
We evaluated each combination of the data imputation techniques, feature selection techniques, and the different machine learning models individually.

\subsubsection{Experimental Setup}
\label{sec:experimentSetup}
To avoid information leakage from the testing set (e.g., during data preparation) it is recommended to partition a subset of the data at the beginning of the project and reserve it for the final evaluation of the best-performing model \citep{lones2021avoid}. 
Thus, 30\% of the data was reserved for the final evaluation of the best-performing model.
The remaining 70\% of the data were split into 10 equal parts using 10-fold cross-validation with stratification on the classes to answer our research questions.
By using stratified cross-validation, we are able to maintain the balance of the data in terms of the distribution of classes within the data, ensuring that each fold is representative of the overall dataset.
To optimize the hyperparameters, each iteration of the cross-validation was split into an additional 10-fold cross-validation.


\subsection{Evaluation Metrics}
\label{sec:evaluation_metrics}
To evaluate the performance of our mobile gaming application installation prediction models, we used two common evaluation metrics: the receiver operating characteristic (ROC) curve and the precision-recall (PR) curve.
The ROC curve is a plot of the true positive rate against the false positive rate at different classification thresholds, and the area under the ROC curve (AUC-ROC) is a measure of the model's ability to discriminate between positive and negative classes.
However, in imbalanced classification tasks such as in this study, where there are significantly more negative examples than positive ones, the AUC-ROC may not be an accurate measure of performance. For this reason, we also computed the area under the precision-recall curve (AUPRC), which is a more suitable metric for imbalanced classification tasks.

The AUPRC of a random estimator can be calculated by dividing the number of positive examples by the total number of examples. Interested readers may wish to read more about these metrics and the relationship between them in \citet{davis2006relationship} and \citet{saito2015precision}.


 
\newpage

\section{Results}
\label{sec:results}
In this section, we present the quantitative results of the compared methods on the Bigabid dataset.

%
\begin{figure}[!h]
    \centering
    \includegraphics[width=0.8\linewidth]{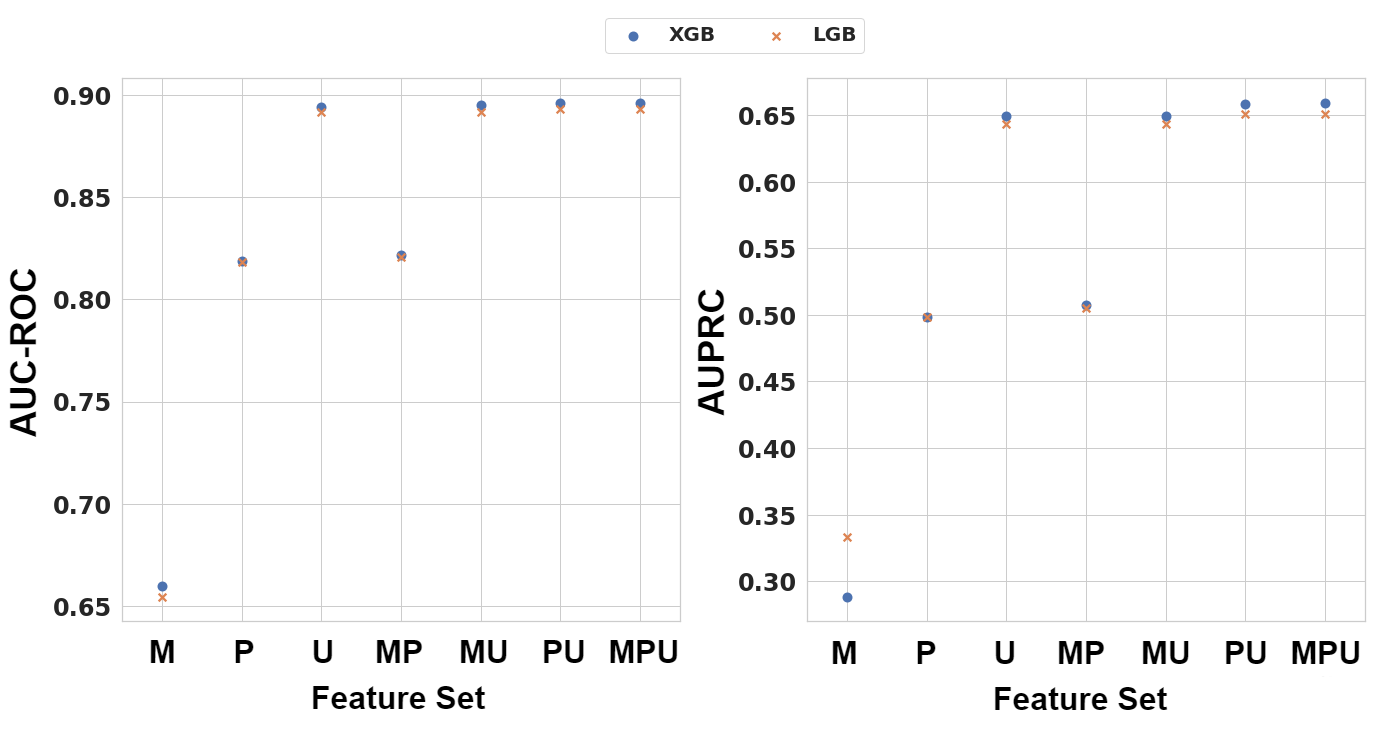}
    \caption{The performance of a model can be satisfactory while still protecting the privacy of the users.}
    \label{fig:feature_Sets}
\end{figure}
%


We tested the performance of seven features combination sets (Meta Features [M], Publisher-Related [P], User-Related Features [U], Meta and Publisher Features [MP], Meta and User Features [MU], Publisher and User Features [PU], and Meta, User, and Publisher Features [MPU]) with two classifiers (XGBoost [XGB] and LightGBM [LGB]) to answer research questions \ref{itm:features_set} and \ref{itm:model}.

Figure \ref{fig:feature_Sets} presents the combination of the feature sets results versus the two classifiers.
Figure \ref{fig:feature_Sets} shows that including the user-related features (U, MU, PU, MPU) in the models leads to better performance in both AUC-ROC and AUPRC.
In contrast, using the meta-features alone (M) results in poor performance of the models.
However, the results of using publisher-related features alone (P) suggest that it is possible to train a model that protects user privacy while still achieving quite good performance.
Overall, both classifiers (XGB and LGB) achieved quite similar performance, but XGB achieved a bit better AUC-ROC and AUPRC results.

\begin{figure}[!h]
    \centering
    \includegraphics[width=\linewidth]{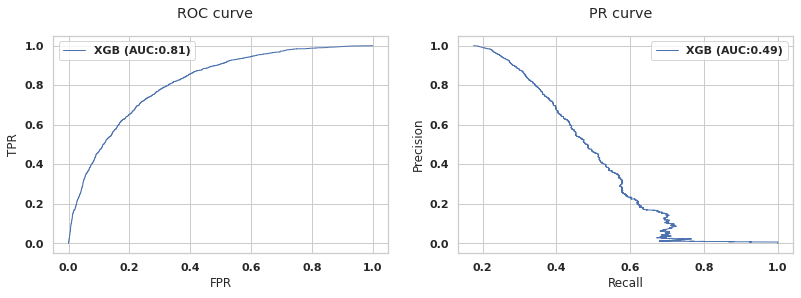}
    \caption{ROC and PR curves of the XGB model with the publisher-related features.}
    \label{fig:curves}
\end{figure}

\newpage

Next, we used XGB, the best-performing model, with the publisher-related features and learned a privacy-aware model using 70\% of the data, and evaluated the model on the 30\% of that data that was reserved for the final evaluation of the best-performing model (Sec. \ref{sec:experimentSetup}).
A privacy-aware model either excludes personal information or uses it only in an aggregated form, which aims at preventing tracking across multiple apps and websites owned by different companies, as well as avoiding the unique identification of individual users or devices. In contrast, a non-privacy-aware model relies on identifiable data such as names, emails, or physical addresses to achieve precise targeting. 

Figure \ref{fig:curves} presents the ROC and PR curves of XGB with the publisher-related features.
As presented in Figure \ref{fig:curves}, in the PR curve, the precision of the model is at 1 at the beginning, since at these decision cut-offs, the model does not make any false alarms.
However, the number of false positives quickly becomes relatively large. For example, to reach a recall of around 0.5, the precision of the model is reduced to below 0.5.







\section{Discussion and Conclusions}
\label{sec:disc_and_conc}
In this study, we investigated the process of predicting a mobile gaming app installation from the perspective of a DSP, while paying attention to user privacy. 
This is important because current laws may force companies to declare that they are collecting data and might even give users the option to opt out of such data collection, which could threaten companies' business models in digital advertising that are dependent on the collection and use of user-level data.
We aimed to predict whether a user will install an app after being presented with an impression, given that the DSP has won the auction in an RTB process.

We found that learning a user privacy-aware mobile gaming app installation prediction model leads to an AUC-ROC of 81\% and an AUPRC of 49\%, while using a non-privacy-aware model leads to an AUC-ROC of 89\% and AUPRC of 66\%.
This suggests that it is possible to train a model that protects user privacy while still achieving good performance.

As a commercial company, Bigabid has limitations on the data that can be publicly shared due to legal requirements and the need to protect the company's customers and confidential business information.
This may affect the reproducibility of our study by other researchers.
Moreover, the Bigabid data are limited to the auctions that Bigabid was successful in winning. 
Thus, this is not the exact distribution in the real world, as there can be auctions with outliers utility estimation.
To accurately learn and evaluate the real-world distribution, we would need to have all labels and predict from the perspective of the auction organizer.
Additionally, the number of slots that were offered and the percentage of auctions that Bigbid won or lost cannot be disclosed.

In our future work, we aim to increase the space of hyperparameters considered during the optimization of our models. 
Ensemble learning, imputation, and feature selection can also improve model performance. 
Additionally, we plan to further investigate the trade-off between privacy preservation and model performance and explore ways to improve the performance of privacy-aware models.

\subsubsection*{Conflict of interest}
The authors declare that they have no conflict of interest.

\newpage


\nocite{*} 


\end{document}